\documentclass[balance,upint,subscriptcorrection,varvw,mathalfa=cal=boondoxo,pdf-a,nofoot]{asmeconf2}

\titleformat{\section}{\mathversion{sansbold}\bfseries\sffamily\raggedright}{\thesection .}{0.5em}{}

\usepackage[nomain,acronym]{glossaries}
\newacronym{moea}{MOEA}{multiobjective evolutionary algorithm}
\newacronym{ea}{EA}{evolutionary algorithm}
\newacronym{fem}{FEM}{finite-element method}
\newacronym{mop}{MOP}{multiobjective optimization problem}
\newacronym{cmop}{CMOP}{constrained multiobjective optimization problem}
\newacronym{moo}{MOO}{multiobjective optimization}
\newacronym{sop}{SOP}{single-objective problem}
\newacronym{ga}{GA}{genetic algorithm}
\newacronym[longplural={evolution strategies}]{es}{ES}{evolution strategy}
\newacronym{de}{DE}{differential evolution}
\newacronym{dm}{DM}{decision maker}
\newacronym{mlo}{MLO}{multi-level optimization}
\newacronym{cfd}{CFD}{computational fluid dynamics}
\newacronym{nfl}{NFL}{No Free Lunch}
\newacronym{cma}{CMA}{covariance matrix adaptation}
\newacronym{modact}{MODAct}{Multi-Objective Design of Actuators}
\newacronym{pf}{PF}{Pareto front}
\newacronym{ps}{PS}{Pareto set}
\newacronym[description={feasability ratio, see Definition~\ref{def:fsr}}]{fsr}{\text{FsR}}{feasability ratio}
\newacronym[description={ratio of feasible boundary crossing, see Definition~\ref{def:rfbx}}]{rfbx}{\text{RFB}\ensuremath{_\times}}{ratio of feasible boundary crossing}
\newacronym[description={normalized ratio of feasible boundary crossing, see Eq.~\eqref{eq:nrfbx}}]{nrfbx}{\text{nRFB}\ensuremath{_\times}}{normalized ratio of feasible boundary crossing}
\newacronym{fvc}{FVC}{fitness violation correlation}
\newacronym{piiz}{PiIZ}{proportion in ideal zone}
\newacronym[description={Pareto front distance to cloud, see Definition~\ref{def:pfd}}]{pfd}{\text{PFd}}{Pareto front distance to cloud}
\newacronym[description={Pareto front cv, see Definition~\ref{def:pfcv}}]{pfcv}{\text{PFcv}}{Pareto front cv}
\newacronym{iz}{IZ}{ideal zone}
\newacronym{iqr}{IQR}{interquartile range}
\newacronym{uea}{UEA}{unbounded external archive}
\newacronym{igd}{IGD}{inverted generational distance}
\newacronym{igdp}{\text{IGD}\ensuremath{^+}}{modified inverted generational distance}
\newacronym{ipmq}{IPMQ}{Interprofessional Project Management Questionnaire}
\newacronym{cdp}{CDP}{constrained dominance principle}
\newacronym[longplural={constraint handling strategies}]{chs}{CHS}{constraint handling strategy}
\newacronym{ai}{AI}{\textit{artificial intelligence}}
\newacronym{ml}{ML}{\textit{machine learning}}
\newacronym{sr}{SR}{stochastic ranking}
\newacronym{omod}{OMOD}{Online MODelling platform}
\newacronym{asme}{ASME}{American Society of Mechanical Engineering}
\newacronym{dac}{DAC}{Design Automation Conference}
\newacronym{gan}{GAN}{generative adversarial network}
\newacronym{vae}{VAE}{variational autoencoder}
\newacronym{cds}{CDS}{computational design synthesis}
\newacronym{pbl}{PBL}{problem-based learning}
\newacronym{cdio}{CDIO}{conceive-design-implement-operate}
\newacronym{pwm}{PWM}{pulse-width modulation}
\newacronym{hvac}{HVAC}{heating, ventilation, and air conditioning}
\newacronym{cad}{CAD}{computer-aided design}
\newacronym{sbx}{SBX}{simulated binary crossover}
\newacronym{dof}{DoF}{degrees of freedom}
\newacronym{sd}{SD}{standard deviation}
\newacronym{ode}{ODE}{ordinary differential equation}
\newacronym{doe}{DoE}{design of experiments}
\newacronym{lhs}{LHS}{latin hypercube sampling}
\newacronym{rmse}{RMSE}{root-mean-square error}
\newacronym{llm}{LLM}{large language model}
\newacronym{pdf}{PDF}{probability density function}
\newacronym{ood}{OOD}{out-of-distribution}
\newacronym{dpp}{DPP}{determinantal point process}
\newacronym{pca}{PCA}{principle component analysis}
\newacronym{tsne}{t-SNE}{t-distributed stochastic neighbor embedding}
\newacronym{anova}{ANOVA}{analysis of variance}

\usepackage[inline]{enumitem}

\graphicspath{{figures/}}

\usepackage{siunitx}

\usepackage{tabularx}

\usepackage[normalem]{ulem}

\newcommand{\vect}[1]{\bm{#1}}
\usepackage{float}

\hypersetup{%
	pdfauthor={Cyril Picard},
	pdftitle={DATED: Guidelines for Creating Synthetic Datasets for Engineering Design Applications},
}

\begin{document}




\title{DATED: Guidelines for Creating Synthetic Datasets for Engineering Design Applications} 

 
%
%
%

\SetAuthors{%
	Cyril~Picard\affil{1}\CorrespondingAuthor{cyrilp@mit.edu}, 
    J\"urg Schiffmann\affil{2},
	Faez~Ahmed\affil{1}
	}

\SetAffiliation{1}{Massachusetts Institute of Technology, Cambridge, MA}
\SetAffiliation{2}{\'Ecole polytechnique f\'ed\'erale de Lausanne (EPFL), Lausanne,  Switzerland}


\maketitle

\begin{abstract}
Exploiting the recent advancements in artificial intelligence, showcased by ChatGPT and DALL-E, in real-world applications necessitates vast, domain-specific, and publicly accessible datasets. Unfortunately, the scarcity of such datasets poses a significant challenge for researchers aiming to apply these breakthroughs in engineering design. Synthetic datasets emerge as a viable alternative. However, practitioners are often uncertain about generating high-quality datasets that accurately represent real-world data and are suitable for the intended downstream applications. This study aims to fill this knowledge gap by proposing comprehensive guidelines for generating, annotating, and validating synthetic datasets. The trade-offs and methods associated with each of these aspects are elaborated upon. Further, the practical implications of these guidelines are illustrated through the creation of a turbo-compressors dataset. The study underscores the importance of thoughtful sampling methods to ensure the appropriate size, diversity, utility, and realism of a dataset. It also highlights that design diversity does not equate to performance diversity or realism. By employing test sets that represent uniform, real, or task-specific samples, the influence of sample size and sampling strategy is scrutinized. Overall, this paper offers valuable insights for researchers intending to create and publish synthetic datasets for engineering design, thereby paving the way for more effective applications of AI advancements in the field. The code and data for the dataset and methods are made publicly accessible at \url{https://github.com/cyrilpic/radcomp}.
\end{abstract}







\section{Introduction}
Recently popular \gls{ai} tools like ChatGPT and DALL-E have given a taste of ``intelligence'' to the general public due to their significant impact on a variety of fields, including art, design, and entertainment. In the field of natural language processing, these models can be used to generate realistic text, which has the potential to be used in a wide range of applications, including chatbots, automated content creation, and even generating news articles. In art, these models allow artists to generate new images, videos, and music. This has left many engineers wondering how it will impact their field.

The leap of machine learning in general and of \glspl{llm} in particular was made possible by a combination of new model architectures and a significant increase in model size (about 175 billion parameters for GPT-3~\cite{brown_language_2020}) associated with the availability of very large datasets. Indeed, \glspl{llm} have demonstrated that machine learning becomes capable of solving new tasks with upward scaling of computation (FLOPs), model size (number of parameters), and data size~\cite{wei_emergent_2022}. An example of such \textit{emergent behavior} is the ability of generative models to draw correct and legible text on images~\cite{yu_scaling_2022}. Beyond these examples, though, data size is often cited as a bottleneck~\cite{shani_lean_2023}.


In engineering design, generative \gls{ai} and \glspl{llm} have already been applied to a variety of tasks, e.g., aircraft shape design~\cite{shu_3d_2019},  linkage mechanism generation~\cite{nobari_links_2022}, and concept sketch analysis~\cite{song_attention-enhanced_2023}. In 2019, researchers urged the community to release more datasets, including multi-modal ones~\cite{panchal_special_2019}. Yet a 2022 review still highlights that datasets in the field remain scarce and small~\cite{regenwetter_deep_2022}. While very large datasets are being published, e.g., 100 million sized LINKS dataset ~\cite{nobari_links_2022}, they remain rare and not always easily accessible. This impacts the reproducibility and comparability of published \gls{ml} methods for engineering design. It also impairs the field's ability to benefit from the advances in \gls{ai} and experience its own \textit{emergent behavior} moments.



Data can generally be categorized as either synthetic or real-world data. The latter is generated by real-world events, e.g., bank transactions, images posted on public web pages, texts written on Wikipedia, patients undergoing X-ray imaging, or autonomous cars driving on the streets. The related challenges include the logistics of collecting that data, processing it into a shared format, assessing its quality and biases, and labeling it~\cite{shani_lean_2023}. The milk frother dataset~\cite{song_attention-enhanced_2023} comprising of sketches made by students,\footnote{\url{https://sites.psu.edu/creativitymetrics/2018/07/18/milkfrother/}} or the BIKED dataset comprising of user-uploaded bike designs~\cite{regenwetter_biked_2021} are examples of real-world datasets for engineering design.
In contrast, synthetic data is typically purpose-driven and generated \textit{artificially}, such as by exhaustively enumerating all design options (e.g., periodic cellular structures~\cite{lumpe_exploring_2021}), by sampling from a design space (e.g.,~\cite{nobari_links_2022}), or by augmenting existing designs or real datasets through perturbations (e.g., ship hulls~\cite{wang_three-dimensional_2022}).

Real datasets in engineering design are unlikely to fulfill the need for large and available datasets for \gls{ml} research. They tend to be small (e.g., about 1000 sketches, or 4500 bike designs). Further, these designs tend to be clustered in specific areas of the design space, limiting the capacity for design exploration. Indeed, existing real-world designs are often deliberately of good quality. From an optimization lens, they can be considered optimal and thus near the edge of the design space~\cite{papalambros_principles_2017}, or they are the result of iterative refinement and share attributes with their parents. Finally, while larger datasets do exist, they are closely guarded by the industry, i.e., not openly available at scale.

Conversely, synthetic data could play a central role in addressing the shortage of large datasets through its capacity to be generated at scale. To quantify the performance of synthetic designs, one could leverage a plethora of tools developed over the past decades in different fields that enable performance quantification for a growing range of engineering designs and systems. However, there are two questions that arise: how should one collect or sample a dataset, and how should one quantify if that dataset is useful or not? It is important to note that not all datasets are equally useful for design and \gls{ml} research. For instance, the ship hull dataset contains 300 shapes resulting from data augmentation methods applied to a single existing hull design. It is therefore limited in both size and scope to a narrow domain of single-hull variants. Conversely, LINKS contains 100 million one-degree-of-freedom linkage mechanisms and more than a billion coupler curves, but as the authors noted, most of the resulting coupler curves are arcs and circles, which are of limited practical interest. 

Creating a good dataset is thus not a trivial task and can be subject to conflicting objectives, yet comprehensive methods to support researchers in this process are not discussed in the literature.

To address some of these shortcomings, in this paper we:

\begin{enumerate}
    \item Provide guidelines for researchers and practitioners on factors to consider while collecting synthetic data for machine learning applications,
    \item Illustrate common issues with a case study of turbo-compressor dataset generation, and
    \item Release the code and data used in the case study for researchers to build upon.
\end{enumerate}

The guidelines cover three key areas of research: (1) design representation and data generation, (2) modeling and simulation, and (3) validation and verification. In the first area, researchers must carefully select the source data and properly characterize its properties. They must select the parameters they want to sample, and the sampling method, and understand the different objectives they need to consider while sampling.
In the second area, researchers must choose appropriate modeling and simulation techniques to generate and characterize synthetic data that realistically represent the source data. In the third area, researchers must validate and verify the synthetic dataset to ensure that it accurately reflects the source data or proves useful for certain data-driven \gls{ml} applications. Finally, while not the scope of this paper, researchers must also properly document and share the synthetic dataset to enable future reuse and collaboration (refer to for example~\cite{gebru_datasheets_2021}).

\section{Data Representation and Selection}\label{sec:selection}

\begin{figure}
    \centering
    \includegraphics{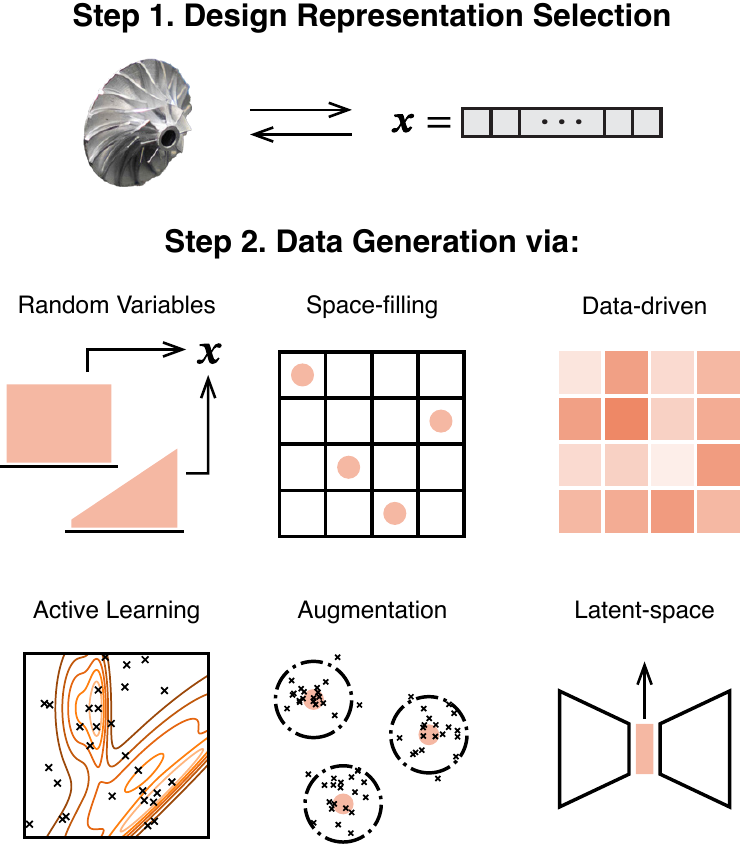}
    \caption{Overview of the important steps for data selection and the general data generation approaches.}
    \label{fig:sum_sampling}
\end{figure}

The first step in creating a dataset is to define what the data will be. This requires answering (i) how the data is represented, and (ii) how the individual data points are generated. The key topics discussed in this section are visually summarized in Fig.~\ref{fig:sum_sampling}.

\subsection{Representation and Feature Selection}

This study focuses on tabular data, which are common in engineering. In tabular design data, each row represents a design variant, and each column is a feature or a label. For example, the UIUC airfoil dataset\footnote{\url{https://m-selig.ae.illinois.edu/ads/coord_database.html}} contains $N=1600$ real-world airfoils that form a table with about $N$ rows, 384 features---the x,y-coordinates of 192 surface points---and one label---the associated lift-to-drag ratio. Formally, a design vector or the vector of features denoted as $\vect{x}$ is the numerical representation of a design within $\mathbb{R}^d$, Eq.~\eqref{eq:x}, where $d$ is called the dimensionality of the space.\footnote{Note: some features can be integers or encoded categories.} The collection of all design vectors forms the design space $\mathcal{D}$, Eq.~\eqref{eq:design_space}, from which new design samples will be drawn in the next step. Labels refer to information that can be derived given a design vector, such as text descriptions, graphical representations, ranks, performance measures, or constraints. The process of gathering labels, i.e., data annotation, is discussed in Section~\ref{sec:modeling}.

\begin{equation}\label{eq:x}
\text{design}_i \longleftrightarrow \vect{x}_i = [x_1, x_2, \dots, x_d] \in \mathbb{R}^d
\end{equation}
\begin{equation}\label{eq:design_space}
    \mathcal{D} = \{\vect{x}_1, \vect{x}_2, \dots \} \subset \mathbb{R}^d
\end{equation}

As in the case of the airfoil dataset, features are often related to parameters in a parametric design framework but they need not be strictly related to geometrical aspects of design. Features can represent colors, materials, or operating conditions. In the context of dataset creation, it is obviously central to carefully choose the link between a design (an airfoil) and its design vector (the x-y coordinates of control points). While that decision is highly context-dependent, the selection of features should in general be:
\begin{enumerate}
    \item \textbf{Compatible} with any existing data and with the relevant models and simulations;
    \item \textbf{Complete} ensuring that all necessary design characteristics are captured;
    \item \textbf{Compact} to reduce feature collinearities and improve processing efficiency;
    \item \textbf{General} enough to enable reuse in different applications and amortize the cost of creating a dataset.
\end{enumerate}

The second and third points should not be seen as directly contradictory. If features correspond to a parameterized design, for example, it may be possible to choose a more compact parameterization without affecting completeness. In such cases, the more compact version should be preferred. Otherwise, if uncertain about a feature, it is better to include it and not use it rather than having to add a feature later. Methods exist to quantify the importance of features in a model, which can aid in the selection process once the dataset exists. Some examples are provided in Section~\ref{sec:validation}.  Finally, it is important to create a dataset that has broader applications, which may include multi-modal learning or using a representation that can generate images or meshes for a design.

The design space associated with a representation may be bounded---i.e., each feature is contained within a range and the resulting space can be defined as in Eq.~\eqref{eq:d_bounds}.
\begin{equation}
\mathcal{D} = \{ \vect{x} \in \mathbb{R}^d | x_i^{(L)} \leq x_i \leq x_i^{(U)} \ \forall \ i=1,2,\dots,d \}\label{eq:d_bounds}
\end{equation}
where \(\vect{x}^{(L)}\) and \(\vect{x}^{(U)}\) are the lower and upper bound vectors respectively. In the simplest case, all bounds are independent, but it is also quite common for the bounds to depend upon others. Defining dependent bounds is particularly interesting if many variable combinations yield ``nonsense'' designs, since this may otherwise make the generation of meaningful designs harder. This is quite common in mechanical engineering where bounds are given on ratios (e.g., the length-to-diameter ratio of holes should be in a given range). Note, however, that this distorts the space, and should be kept in mind when sampling (e.g., very long holes in absolute terms are harder to get at random since it depends on the diameter).
In addition, in some cases, upper and/or lower bounds cannot be defined and the design space is then called unbounded. Again, this results in special challenges that need to be considered in the data generation part.

\subsection{Data Generation}
Once the representation and the design space are defined, designs within it can be generated. This section discusses the objectives of the process, the question of the dataset size, as well as the many methods that can generate the actual data.

\subsubsection*{Objectives}
Before discussing the hows, it is important to know what are the objectives of the datasets. For example, in metamaterial dataset design, the objective may be to get metamaterials that cover a preset property range, or the objective could be to differentiate between manufacturable and unmanufacturable metamaterials. 
In general terms, the goal of data generation is to collect a set of \textit{relevant} points, where \textit{relevant} is context-dependent. It is often important to ensure that the dataset has sufficient diversity. Data-driven methods usually have degraded performance for \gls{ood} data. So, having good coverage of the space is important to improve the generalizability of machine learning models trained on those datasets. Likewise, a machine learning model's accuracy is often highest in regions with a high density of points.

\subsubsection*{Dataset Size}
How many data points are needed? The answer first depends on the dimensionality $d$. If one divides, for example, all features into three values, one already needs $3^d$ points. For thirteen features ($d=13$), that is more than 1.5 million points. This phenomenon is referred to as the curse of dimensionality. It becomes very clear that achieving a high density of points across the whole design space is impossible. There is therefore a trade-off between space coverage (for design exploration) and high local density in regions of interest. Still, in general, more data is preferred for improving the performance of machine learning methods~\cite{sorscher_beyond_2022}. If needed, it is always possible to downsample for specific tasks. The bottleneck for dataset size, however, is usually the required budget for data annotation (some advice will be provided in Section~\ref{sec:modeling}).

\subsubsection*{Sampling}
In the context of synthetic dataset generation from parameters in given ranges, sampling refers to the process of randomly selecting values for each parameter from their respective ranges. This approach is commonly used to create synthetic datasets for machine learning or other types of modeling where real-world data may be limited or not available.
As a reminder, the parameters can represent various characteristics or features of the data, such as numerical values, categorical variables, or distributions. For example, if we are generating a synthetic dataset for a regression model for bicycle structural performance, the parameters might include continuous variables such as the length of the top tube or bottom tube, each with a specified range of values or categorical options.

In this work, sampling methods have been grouped into six categories: (1) random sampling, (2) space-filling sampling, (3) data-driven sampling, (4) active-learning sampling, (5) data augmentation, and (6) latent-space sampling, summarized graphically in Fig.~\ref{fig:sum_sampling}.

Below, we discuss these six different sampling methods, highlighting their advantages and disadvantages.

\subsubsection*{Random Sampling}
In this approach, each sample is drawn from a random distribution. The uniform distribution is the most commonly used in this setting. When the design space is bounded, drawing uniform sampling is fast and if enough points are considered, it will generally yield a good overall coverage. Not surprisingly, many datasets are collected that way. It is also possible to go beyond uniform distributions to achieve more tailored behaviors. For example, many distributions are unbounded, e.g., normal distributions, and can thus also be applied to unbounded spaces. For features that cover several orders of magnitudes, uniform sampling will generate most samples on the largest scale.  If for example, a feature corresponds to a pressure that can span from \si{\kilo\pascal} to \si{\mega\pascal}, most samples will be in the order of \si{\mega\pascal}. Power distributions or uniform distributions in the logarithmic space can be used to ensure that more samples are spread more evenly within each scale. For categorical features, values can be selected randomly from a list of given categories.
While random sampling is fast to run, it may not always ensure that the design or feature space has good coverage. It may also ignore relationships between features, and a significant number of samples may get wasted, as they may not lead to any feasible design.

\subsubsection*{Space-filling Sampling}
Space-filling sampling approaches aim to provide good coverage over the design space. Classical approaches (e.g., factorial design) put a grid with fixed partitions onto the space and allocate a point to each node. The resulting number of points can be extremely large due to the curse of dimensionality ($k^d$) detailed previously. More recent approaches focus on covering the space best with a fixed but choosable number of points. Examples include \gls{lhs}~\cite{mckay_comparison_2000} or Sobol sequences~\cite{joe_constructing_2008}. They work by partitioning the space into smaller regions but follow point placement rules. In particular, Sobol sequences can be advantageous since they guarantee certain coverage properties while being computationally efficient~\cite{kucherenko_exploring_2015}. Also, the code to generate such sequences is readily available in major \gls{ml} frameworks. However, the number of samples is typically restricted to powers of two, and adding points sequentially while maintaining the space-filling properties can be challenging,

\subsubsection*{Data-driven Sampling}
In data-driven sampling, the sample is selected based on the characteristics of the data, such as its distribution, variability, or patterns. Sometimes similar in intent to space-filling approaches, data-driven sampling methods do however not manipulate space, but data. As such, some data must already exist or have been generated by another method.
One common approach to data-driven sampling is cluster sampling, where the data is first partitioned into clusters based on similarity, and then a sample is selected from each cluster. This approach can be useful for identifying patterns or trends within the data, as well as for reducing the computational burden of analyzing large datasets.
Also popular, \glspl{dpp}, which given a similarity matrix, model the likelihood of selecting a diverse subset~\cite{kulesza_determinantal_2012}. The definition of similarity controls the selection preference. Multiple preferences can be blended together with a parameter $w$ using Eq.~\eqref{eq:mix}. An approach successfully applied in~\cite{chan_metaset_2020} to have samples diverse in the design and in the performance space.

\begin{equation}\label{eq:mix}
L = (1-w) L_1 + w L_2
\end{equation}

Data-driven sampling can be advantageous over other sampling methods because it allows for a more nuanced and flexible approach to selecting the sample. However, it can also lead to 
 bias or inaccurate conclusions if the data is not representative of the larger design space. 

\subsubsection*{Active-learning Sampling} In an active learning context, a model of the data is built iteratively. New sample points are selected by optimizing a utility (or acquisition) function over this model. To incorporate performance information, new samples are usually annotated as they are sampled. Bayesian optimization is a popular active-learning flavor, but other approaches are possible. For example in t-METASET, a \gls{dpp}-based active-learning sampling framework is presented that considers design and performance diversity in general while being able to tailor the distribution of points according to task-specific needs~\cite{lee_t-metaset_2022-1}. The availability of labels and the diversity of possible utility functions make it a very powerful tool. For example, when large regions of the design space yield invalid/infeasible designs---i.e., designs that violate some constraints---active learning can discover the feasibility boundaries and allocate samples more effectively~\cite{bryan_active_2005}. The added information comes obviously at a price. Recent advances have tried to alleviate those issues by enabling batches of samples to be selected~\cite{wilson_maximizing_2018} and running the active-learning model on GPU~\cite{balandat_botorch_2020}.

\subsubsection*{Data Augmentation}
When a small dataset, for example of existing designs, is already available, data augmentation techniques can be applied to generate more similar samples.
In data augmentation-based sampling, the original data is used as a starting point, and various transformations or augmentations are applied to generate new samples. For example, in the context of image data, augmentations can include flipping the image, cropping a part of the image to create a copy, rotating the image, changing the brightness or contrast, or adding noise to the image.
In engineering design, augmentation is often achieved by applying a small perturbation to the designs, e.g.,~\cite{wang_three-dimensional_2022}. In practice, this can be done by sampling from a normal distribution whose mean is set to the actual design and the variance to a fraction, e.g., 1\%, of the length of the design space. Other approaches include using the mutation and crossover operators from genetic algorithms~\cite{hadka_borg:_2013}.
By generating new samples through data augmentation, the size of the training dataset can be increased without the need for additional data or annotation collection. However, augmentations should often be carefully collected using domain knowledge.

\subsubsection*{Latent-space Sampling}
With the development of generative \gls{ai}, researchers have started using \glspl{vae}, \glspl{gan} or other deep generative models to generate new designs by sampling in the latent space~\cite{nobari_pcdgan_2021}. As the latent space is a reduced dimensional space, sampling there reduces issues such as collinearity/correlations between features. In addition, deep generative models can be trained to favor certain properties such as diversity and novelty~\cite{regenwetter_deep_2022}. They can also be conditioned to generate designs with certain performances~\cite{nobari_pcdgan_2021}. The obvious challenge to this approach is that a deep generative model with the desired properties needs to exist or be trained first.

\section{Modeling and Simulation}\label{sec:modeling}
While covering all aspects of modeling and simulation is out of scope for this work, there are several aspects and trade-offs that need to be discussed 
in the context of dataset creation.

As previously stated, labels (also referred to as ground truth, tags, metadata, performance metrics, or constraints) can be very different and can include numerical performance measures (e.g., efficiency, power output, drag coefficient), images (e.g., rendering of the design or flow fields), points clouds (e.g., 3D representations), or rankings (e.g., design A is preferred over design B). While obtaining some may require complex finite-element methods, others may be straightforward to calculate (e.g., calculating the cross-sectional area of an airfoil).

When deciding which labels to include and how to obtain them, researchers and practitioners should consider (i) the level of detail of the data representations, (ii) the needed accuracy for the application, and (iii) the available solvers. Here again to promote reuse, as many labels as possible should be considered to have a multitask-ready dataset. Researchers should also consider purposefully including different label types to create rich datasets.

When collecting dataset labels using computational simulations or analytical methods, it is important to carefully select an appropriate model, tune the model parameters, validate the accuracy of the model, optimize computational efficiency, and format the labels in a consistent and usable way.
This process requires careful consideration and validation to ensure that the generated data is reliable and accurate.

If the desired simulations are too expensive for a large dataset, it may be possible to compromise on the accuracy by running a high-fidelity solver for fewer iterations, or by using a reduced-order model instead~\cite{benner_survey_2015,massoudi_robust_2022}.
If high accuracy is nonetheless desired, a multi-fidelity approach might offer a better trade-off. In multi-fidelity modeling, models of different fidelity are available and an active learning strategy searches for the optimal computational budget allocation strategy~\cite{sarkar_multifidelity_2019}.
Alternatively, weak labeling approaches can be considered~\cite{shani_lean_2023}. Weak labeling makes use of proxy labels for the originally desired annotation. For example, a performance ranking procedure can be used in place of an expensive evaluation of the exact performance (e.g.,~\cite{chaudhary_water_2020}). Similarly, weak labeling works best if some samples can be annotated with the original method, and active learning approaches should then be applied to decide which sample to annotate with which method. 

Despite the associated computational cost, we want to emphasize that the value of a dataset comes from its labels. As a counter-example, ShapeNet is a large-scale 3D shape database~\cite{chang_shapenet_2015}, which contains many engineering-relevant designs (amongst others: 4,043 airplane models). However, its use for engineering design is limited since these shapes have no associated performance label. Researchers that want to use for example those airplane models, first need to evaluate their own labels.

Finally, for a high-quality dataset, the results generated in this step should be carefully post-processed to properly label errors and handle unlikely values.

\section{Validation and Verification}\label{sec:validation}

\begin{figure}
    \centering
    \includegraphics{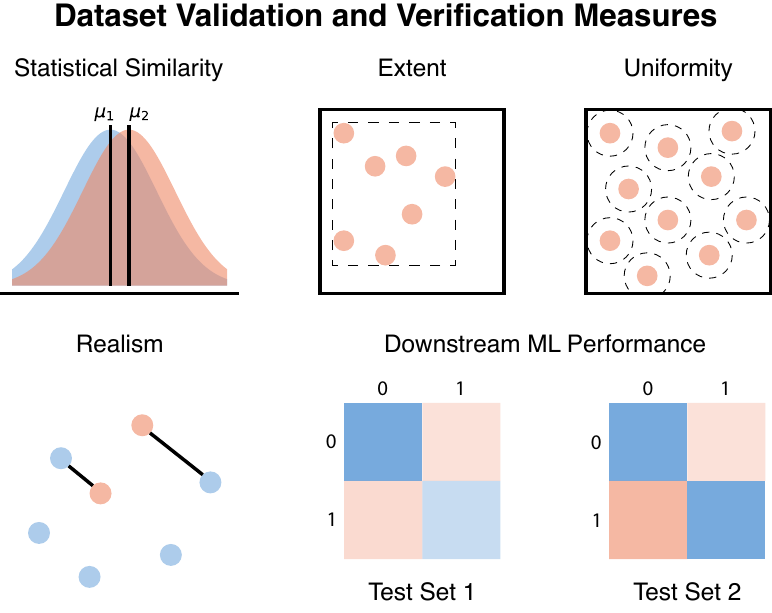}
    \caption{Overview of the approaches that can be used to validate the characteristics of a dataset, and verify the dataset's suitability.}
    \label{fig:sum_validation}
\end{figure}

With the data generated and annotated, there remains an important, and sometimes overlooked, step: validation and verification. The collected data should be accurate, complete, and representative of the real-world system being modeled or observed. It is also important to ensure that the data is relevant and appropriate for the intended use case or application, while also staying free from biases that could impact the results or conclusions of downstream analyses. 
Finally, ensuring that the data is consistent and reliable across different samples, experiments, or observations is important. This can involve comparing the data to known or expected results, or evaluating the data for consistency over time or across different data sources. The goal of validation and verification is to validate that the dataset has the desired characteristics and to verify that those characteristics transfer into good performance on downstream tasks. An illustrative overview of the validation and verification approaches discussed in this section is provided in Fig.~\ref{fig:sum_validation}.

\subsection{Characterization}
This section is all about available methods that can be used to obtain a qualitative and quantitative understanding of a large dataset.
Before diving into the methods themselves, it is important to keep the two following points in mind. First, most methods have underlying assumptions---e.g., linear relationships or normality. Therefore, it is advised to always combine several methods. Second, many characterization methods rely on the existence of a similarity and/or distance measure between two designs. If a context-specific method exists, that method should be preferred. For distance measures, common choices include the Euclidean distance (or $L^2$ distance), the Manhattan distance (or $L^1$ distance), or the Hausdorff distance (which measures the distance between sets of points and would, for example, be appropriate to measure the distance between airfoil profiles). Similarity measures are often built by ``inverting'' a distance measure, for example, with a radial basis function kernel where the similarity $S(i,j)$ between sample $i$ and $j$ can be expressed as a function of their distance $d(i,j)$, $S(i,j) = \exp \left( -0.5 d(i,j)^2 \right)$. Another common, similarity measure is cosine similarity, which is the cosine of the angle between two samples. Lastly, it is worth mentioning that in most measures, multiple features are being summed in some way. So, large differences in scale can lead to biases towards certain features. For this reason, it may be interesting to consider calculating these similarity/distance measures in the normalized or standardized design space, or in a learned embedding (e.g.,~\cite{chan_metaset_2020}). For a broader discussion on metrics, readers can refer to the discussion in~\cite{regenwetter_beyond_2023}, which can be transposed to the context of dataset creation.

\subsubsection*{Statistical Methods}

Typical in \gls{ml}, datasets are characterized using a collection of statistical approaches. For individual or pairs of features/labels, this includes looking at histograms to assess their distribution, quantifying unbalance for categorical data, or calculating descriptive statistics (e.g., mean, variance, median, interquartile range,\dots). The objectives are to (i) validate that the dataset has produced the desired feature distributions, (ii) investigate the resulting label distributions, and (iii) identify extreme values and potentially unrealistic values. The latter is particularly important when using numerical models which may not converge correctly, and involves making sure quantities such as weights or pressures are positive, or that efficiencies are between zero and one.

Further, relations within features or between features and labels can be evaluated, for example, by calculating correlations (linear model: Pearson's $r$ or nonparametric: Spearman's $\rho$) or performing an \gls{anova} or a \gls{pca}. Strongly correlated features can be indicative of collinearities (redundant features) that would cause issues when training \gls{ml} models.  It is however important to note that some degree of correlation may also reflect biases in the sampling. For example, in a dataset of holes, the diameters and the lengths may be correlated because deep holes are more frequently associated with large diameters without it implying collinearity. 
In addition, \gls{anova} or \gls{pca} can provide a first insight into the relations and importance of features with respect to labels.

Clustering methods can also support finding patterns that could be indicative of biases in the data. K-means, for example, is such an algorithm and it will try to identify n groups of equal variance within the data. Clustering can be used in combination with dimensionality reduction techniques, such as \gls{pca} or \gls{tsne}, to visualize in a low-dimensional space high-dimensional data.

\subsubsection*{Diversity} Diversity can be considered both in the design and the performance space. It refers to how well a dataset covers the extent of that space and how uniformly the data points are spaced out. These two components can either be quantified separately or together in a single metric. The extent also called spread, is often measured in terms of bounding volume. This typically means finding the smallest box or sphere that fully encloses the data points and calculating its volume. Depending on the distribution of points, bounding boxes and spheres tend to overestimate the extent. An alternative is to calculate the volume of the convex hull of the data points. The convex hull will usually be much tighter around the points since it is similar to wrapping a plastic film around them. In all cases, larger volumes mean more spread.
Uniformity is typically measured by looking at the distances between neighboring points. In a uniform dataset, these distances will be similar across all points. Whereas if there are differences in distances, points with large distances to nearest neighbors would indicate a low-density area and small distances would indicate a high-density area. Clustering can be used in such cases to identify these different areas.
There are also a few metrics that combine both uniformity and extent. Some common metrics are the Shannon entropy index, or the DPP diversity score~\cite{kulesza_determinantal_2012}. The latter calculates a diversity score based on the eigenvalue decomposition of a pairwise similarity matrix.



\subsubsection*{Realism} In this work, realism refers to the representativeness of the dataset with respect to real-world data. It can be considered in a statistical similarity sense, where realism would mean that the biases of real-world data are also present in the dataset. The Kullback-Leibler divergence is an example of a statistical similarity measure. Further, realism can be understood as a proximity measure. Datasets whose points are close to real data would have a high realism. The difference with respect to the statistical similarity is that the frequency of data points close to real data is less important. Typical set-to-set distance measures include the Hausdorff or the Chamfer distance. Both work by calculating the distance to the closest point from each set to the other, and then either taking the maximum or the average.

\subsection{Measuring Dataset Usefulness}
The final step of the creation of a dataset is to use it with some data-driven model and assess whether its characteristics enable the desired downstream performance.

\subsubsection*{Representativeness of Test Sets}
As data-driven models are optimized towards having good performance on the samples provided for training, their performance should be assessed on data never seen before by the model. This set of data is called a test set. It plays a central role whether comparing the performance of different models or in this context, verifying the suitability of the created dataset.

Most commonly, test sets are created by randomly splitting the data into train and test sets---many \gls{ml} methods also use a validation set, derived from the train set, to monitor the training progress. The resulting test set will most likely have a similar distribution to the training data. While certainly always good to have, it may not be a good predictor of the performance of the model if the conditions are different when the model is deployed. For example, if the goal is to create a surrogate model to be used in an optimization routine, the inputs will rapidly be biased towards designs with high performance.  As such, one may want to have a test set to specifically test the accuracy of a model on high-performing designs.

Test sets are key to the assessment of \gls{ml} models. They should be sufficiently large to reduce random effects and be representative of easier to more complex \gls{ml} tasks. Further, we recommend defining several test sets with different characteristics and providing them along with the data.  That way, they can be indicative of the strength and limitations of various \gls{ml} methods, but also of the dataset itself. As such, in addition to the standard test set similar to the training data, researchers should consider test sets that (i) verify the generalizability in the design and performance space, (ii) assess the realism, and (iii) are representative of the deployed context. For example, the following test sets could cover those objectives:
\begin{itemize}
    \item Similar to the training data (the standard approach);
    \item Diverse in the design space;
    \item Diverse in the performance space;
    \item Similar to real designs;
    \item Composed of designs outside or at the edge of the design space;
    \item Task-specific (e.g., high-performing designs only).
\end{itemize}
In practice,  bespoke data-driven sampling methods can be created to obtain most of these desired goals, while others may require domain knowledge. 

\subsubsection*{Verification}
With the train set and the test sets fully defined, the final task is to actually evaluate different models on various tasks and see how state-of-the-art methods perform on the newly created dataset.
Here as well, multiple models considering multiple tasks should be trained to verify the dataset's suitability. In addition to the common supervised classification and regression tasks, researchers should also apply their dataset to unsupervised autoencoding tasks, or self-supervised or supervised generative tasks.

Once a model is trained, feature importance analysis methods, such as SHAP~\cite{lundberg_unified_2017-1} or permutation approaches~\cite{breiman_random_2001}, can provide the weight of each feature in the predicted outcome. Thus, it becomes possible to loop back to the question of data representation and feature selection and reevaluate if features are missing or should be removed. On this final note, readers are reminded that while the process has been described in a linear fashion, the task of creating a synthetic dataset is often an iterative process.

\section{Case Study: Centrifugal Compressor Design}
\begin{figure}
    \centering
    \includegraphics[width=\linewidth]{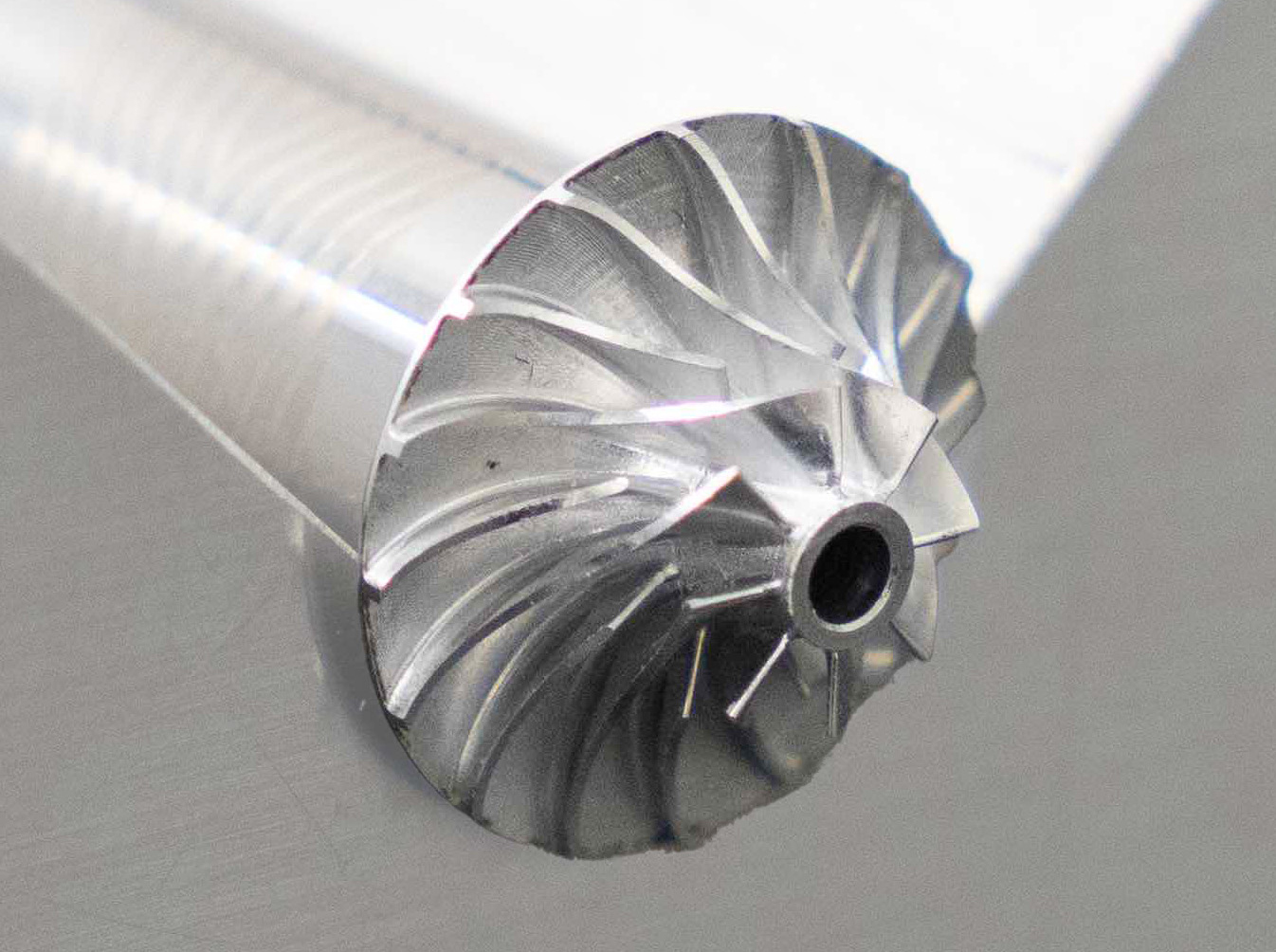}
    \caption{Picture of a small-scale centrifugal compressor supported on gas bearings (by the Laboratory for Applied Mechanical Design at EPFL).}
    \label{fig:comp_photo}
\end{figure}

To illustrate some of the trade-offs and concepts presented in the previous sections, we consider the design of centrifugal compressors, see Fig.~\ref{fig:comp_photo}.

\begin{figure*}[h!t]
    \includegraphics[width=\linewidth]{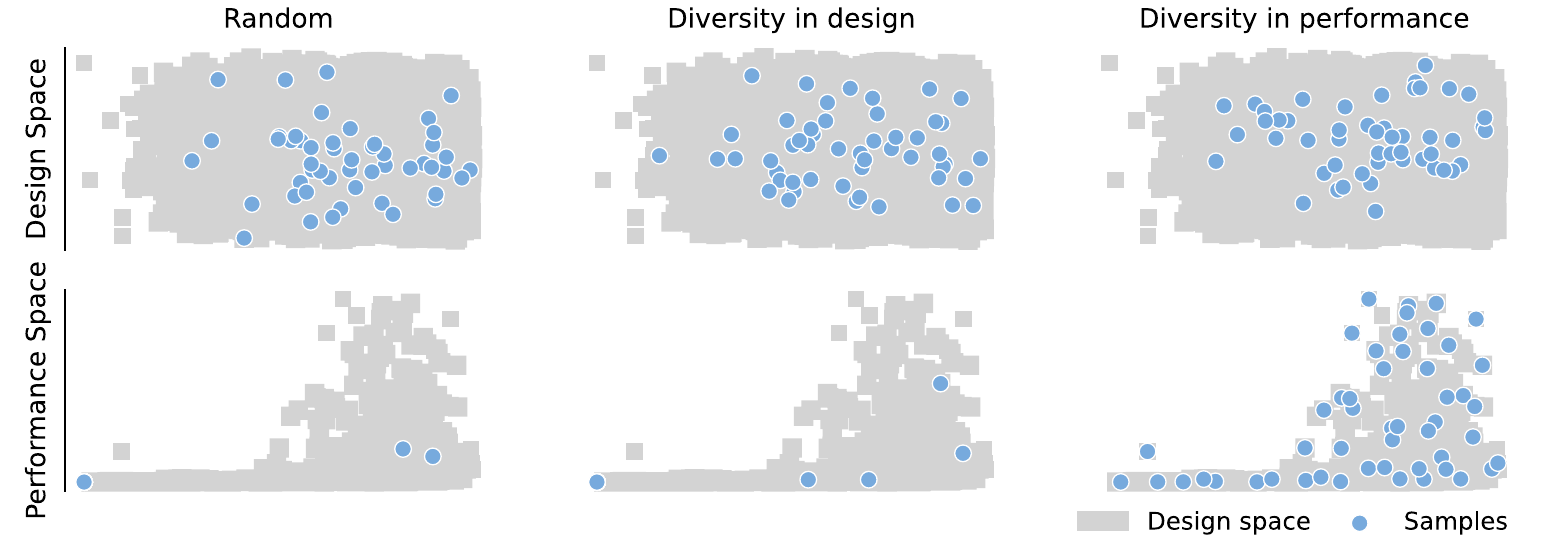}
    \caption{Comparisons of the distribution of 50 data points in the PCA-embedded design space (top) and in the performance space (bottom) selected at random (left), diversely sampled in the design space (middle), and diversely sampled in the performance space.}
    \label{fig:diversity}
\end{figure*}

Centrifugal compressors play an important role in increasing the performance of internal combustion engines. Due to their efficiency and high power density, they are increasingly used for heat pump and refrigeration applications~\cite{javed_small-scale_2016}, or waste-heat recovery~\cite{demierre_modeling_2014}. Their design undergoes a long process that starts by defining key dimensions, such as tip diameter and operating speed, and ends with complex three-dimensional flow analyses. Previous work has shown that data-driven methods can greatly support and speed up the first stages of their design process~\cite{mounier_data-driven_2018,massoudi_robust_2022}. These contributions use a mean-line analysis model validated using experimental data~\cite{schiffmann_design_2010} to evaluate the feasibility and the performance of each design in less than a second. The model has been converted to Python and is publicly released for the first time along with the generated data.\footnote{ \url{https://github.com/cyrilpic/radcomp}}

This case study offers the advantage of being complex enough---it has 21 features and three labels---while still being computationally affordable. Further, we have collected the designs of 14 real-world compressor geometries from the literature~\cite{schiffmann_design_2010,javed_performance_2011, javed_small-scale_2016,de_bellis_validation_2013,meroni_design_2018,olmedo_high-speed_2023}. They form the dataset of \textit{real} designs and their performance is evaluated with the same model as synthetic samples.

\subsection{Methods and Experimental Conditions}
The features correspond to the geometrical parameters relevant in the predesign stage, and to the operating conditions at which each compressor will be evaluated. They are listed along with their range in Table~\ref{tab:compressor_features} in Appendix.  The ranges are set to cover the set of existing designs and follow general recommendations for centrifugal compressors~\cite{baines_fundamentals_2005}.

For the purpose of this work, a total of 22 million samples have been generated: 19 million by random sampling, two million by data augmentation of the 14 real-world compressors, and, in a second step, another million by random sampling of a restricted part of the design space (discussed in the next section). The random number generator routines of \textit{numpy}\footnote{Numpy 1.23.3 and Python 3.9.13} were used for sampling. All features were sampled uniformly within the specified ranges, except for the reduced inlet pressure $P_{r,1}$ which was sampled from a power distribution with $\alpha=5$. The data-augmented designs were obtained by applying a normal noise with $\sigma$ of 1\% of the design space to the 14 real-world compressors.

The samples were then annotated using the aforementioned mean-line analysis model. Given geometrical parameters and operating conditions (inlet pressure and temperature, mass flow, and rotational speed), the model calculates the input and output velocity triangles at the impeller, and the velocity and flow angle at the inducer and the vaneless diffuser. The mean-line flow is corrected by accounting for losses using established correlations. The performance of the compressor is then given by the total-to-total isentropic efficiency and the pressure ratio. In addition, the model outputs a flag indicating whether the compressor is in \textit{working} condition. \textit{Non-working} conditions include condensation, choke, or surge. When \textit{non-working}, no meaningful efficiency or pressure ratio can be calculated, and their values are set to 0 and 1, respectively. The model evaluations were performed on MIT SuperCloud~\cite{reuther_interactive_2018}.

To verify the performance and suitability of the dataset, the classification into \textit{working}/\textit{non-working} is selected as the target \gls{ml} task.
Consequently, the following three test sets are defined:
\begin{enumerate}
    \item Uniform: random samples from within each class from the random dataset (balanced 150,000 samples);
    \item Real: random samples from within each class from the augmented dataset (balanced 40,000 samples);
    \item Specialized: samples obtained by evaluating real designs on a dense grid of operating conditions and by retaining only points that are near the edge of the \textit{working} boundary (balanced ~40,000 samples).
\end{enumerate}
The samples included in a test set are removed from the train sets. In the next sections, the combination of both random sample sets is referred to as the random dataset (19.4 million samples), while the augmented set refers to the samples generated through data augmentation (1.37 million samples).

As \gls{ml} modeling is not the focus of this work, we choose the state-of-the-art AutoGluon's tabular predictor~\cite{erickson_autogluon-tabular_2020}. AutoGluon trains an ensemble of different \gls{ml} models and tunes their hyperparameters to create an aggregated predictor. For all cases, we have used the ``good quality'' preset and used one GPU for training. No time budget was set and the default accuracy metric was set as the tuning criterion.

\subsection{Discussion}

In this section, we will first present some of the characteristics of the obtained dataset and discuss how diverse sampling in different spaces impacts the selected samples. Second, we train a set of classifiers on train sets of different sizes and of different compositions and report the resulting performance on different test sets. Finally, we give an example of feature importance analysis.

\subsubsection*{How to sample diverse points?}
Analyzing the 19 million random samples, one characteristic strikes out immediately: only about 8\% have been labeled as \textit{working}. While the sampling bounds have been set to maximize the number of \textit{working} compressors, it remains that random combinations of compressor geometry and operating conditions result mostly in \textit{non-working} data points. Consequently, this dataset is highly unbalanced, and by extension, training a classifier on that is difficult. Further, it also means that 92\% of the samples have an efficiency and a pressure ratio set at 0 and 1, respectively. Feeding the dataset to a regressor to predict either one will equally not result in a high-performing model. This is a common situation that many researchers have probably encountered, and we discuss some approaches to this problem.

While there are ways to tackle the imbalance at the model level---by adding different weights to each sample for example---there will always be the issue of information unbalance. So instead, the effect of different sampling approaches is considered here. Figure~\ref{fig:diversity} shows the distribution in the design and performance space of points sampled randomly, considering diversity in design and considering diversity in performance. It highlights that having good diversity in design space is not correlated with good diversity in performance space. As a consequence, using a space-filling sampling alone would not avoid the imbalance. Figure~\ref{fig:diversity} also demonstrates that the design diverse sampling has fewer clusters and might be beneficial for covering the design space more effectively with fewer samples. An overall better diversity could be achieved by using a data-driven method that blends in the two objectives, see for example the work in METASET~\cite{chan_metaset_2020}.

\begin{figure}
    \centering
    \includegraphics{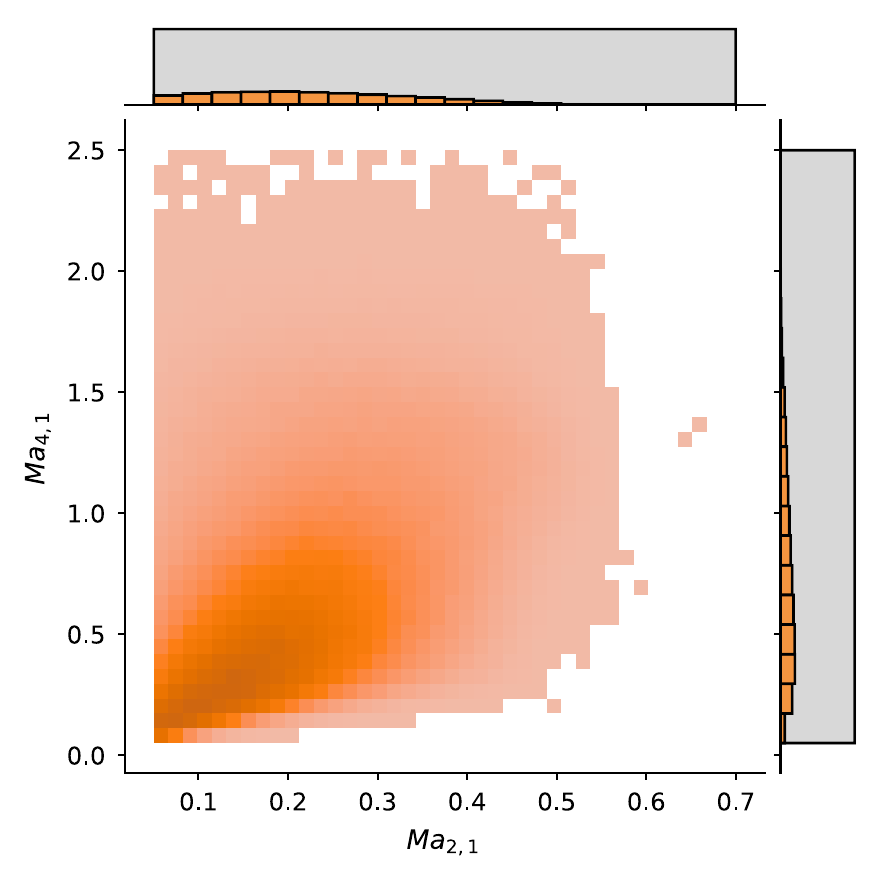}
    \caption{Density plot of working data points in the $(Ma_{2,1}, Ma_{4,1})$ space, with marginal distributions displaying their share with respect to non-working points, highlighting the area where most working samples can be found.}
    \label{fig:adim_map}
\end{figure}

Alternatively, it is sometimes also possible to directly use domain knowledge to infer where more \textit{working} samples can be found. In the case of a compressor, the inlet and impeller tip Mach numbers, $Ma_{2,1}$ and $Ma_{4,1}$ proportional to the mass flow and the rotational speed respectively, are important drivers. Indeed, Fig.~\ref{fig:adim_map} shows the density of \textit{working} samples in the random dataset with respect to these two variables. While \textit{working} samples can be found almost everywhere, there is a high-density area in the lower-left corner. Using that information, a second batch of random samples was drawn over a reduced range with $Ma_{2,1} \in [0.15, 0.25]$ and $Ma_{4,1} \in [0.35, 0.7]$. In that batch, 44\% of the samples are \textit{working}. Figure~\ref{fig:adim_map} also illustrates the kind of characteristics that can be learned by looking at feature distributions.

Overall, it is important to consider coverage over both design and performance space, as a lack of it can lead to degraded \gls{ml} model performance. Visualization methods can help better understand the characteristics of these spaces and inform changes to sampling methods.

\subsubsection*{How to choose the sample size?} 
\begin{figure}
    \includegraphics[width=\linewidth]{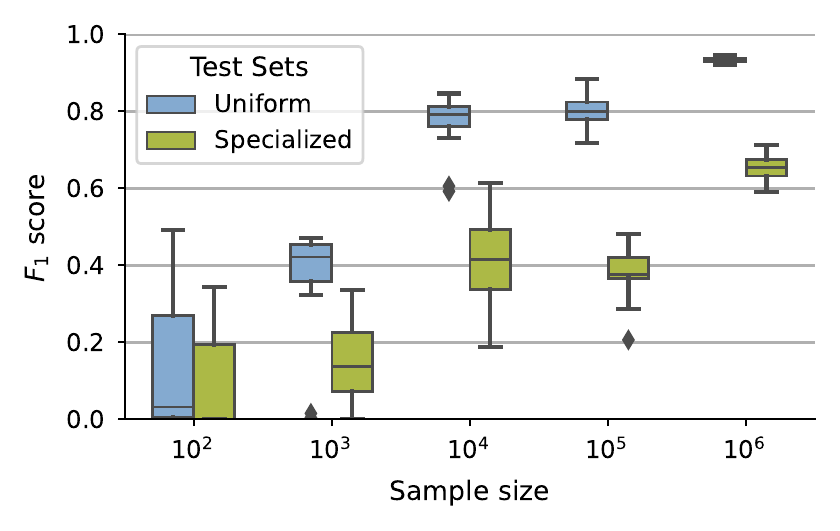}
    \caption{Evolution of the classification performance ($F_1$ score) on the uniform and specialized test sets of AutoGluon models trained with varying sample sizes. For each, ten independent models, are trained with samples drawn at random from the random dataset.}
    \label{fig:samplesize}
\end{figure}

The common intuition in dataset collection is that more data is always better for machine learning performance. In this experiment, we test this intuition. 
We use the trained classifiers to investigate the relationship between classification performance and the train set size. The models are tested with the uniform and the specialized test sets for sample sizes from $10^2$ to $10^6$ drawn from the random dataset. The $F_1$ score \eqref{eq:f1} is used as the performance metric. The results are shown in Fig.~\ref{fig:samplesize}. Sample sizes smaller than $10^4$ lead to high variability and low performance in general. Starting with $10^4$, most trained classifiers have an $F_1$ classification score greater than 0.6. With the largest sample size, very good performance with marginal variations is measured with the uniform test set. It is interesting to note that the performance of models on the uniform test set is always better than the specialized test set. The specialized test set evaluates precisely the edge of the \textit{working} domain, and the performance on it does not reach very good performance levels even with a million samples.

\begin{equation}\label{eq:f1}
    F_1 = \frac{2 \text{ true positives}}{2 \text{ true positives}+\text{false positives} + \text{false negatives}}
\end{equation}

The experiment highlights the importance of having several test sets and ensuring they fit the application. Indeed, if in practice, such a classifier would work with data similar to the uniform test set, then one could conclude that $10^6$ is a sufficiently large dataset and one could even settle with a smaller dataset if memory-constrained. However, the selected uniform training data sampling method does not seem a good fit in a scenario where the data are similar to the specialized test set.

In general, it is a good practice to see how the performance of a model evolves as the dataset is reduced.

\begin{figure*}[htbp]
    \includegraphics[width=\linewidth]{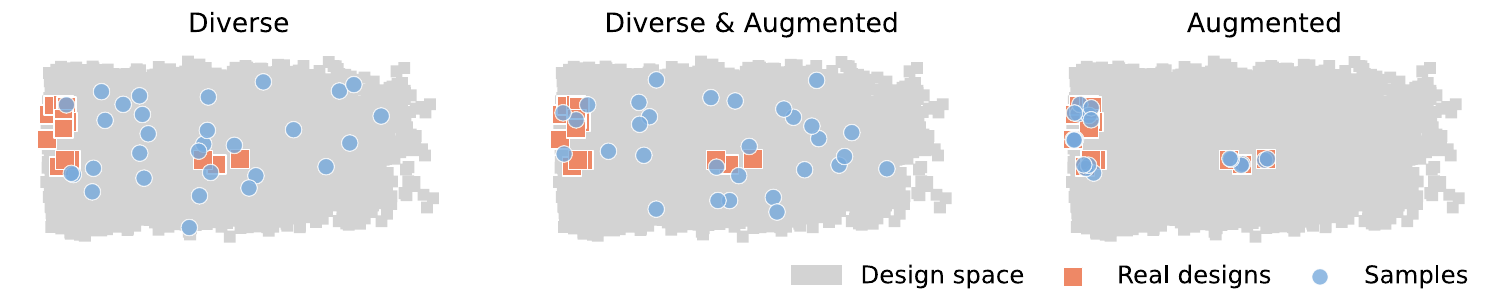}
    \caption{Comparisons within the PCA-embedded geometrical parameter space of the distribution of real designs against 30 data points sampled considering diversity (left) diversity and augmentation (middle), and augmentation only (right).}
    \label{fig:realism}
\end{figure*}

\begin{figure}[H]
    \includegraphics[width=\linewidth]{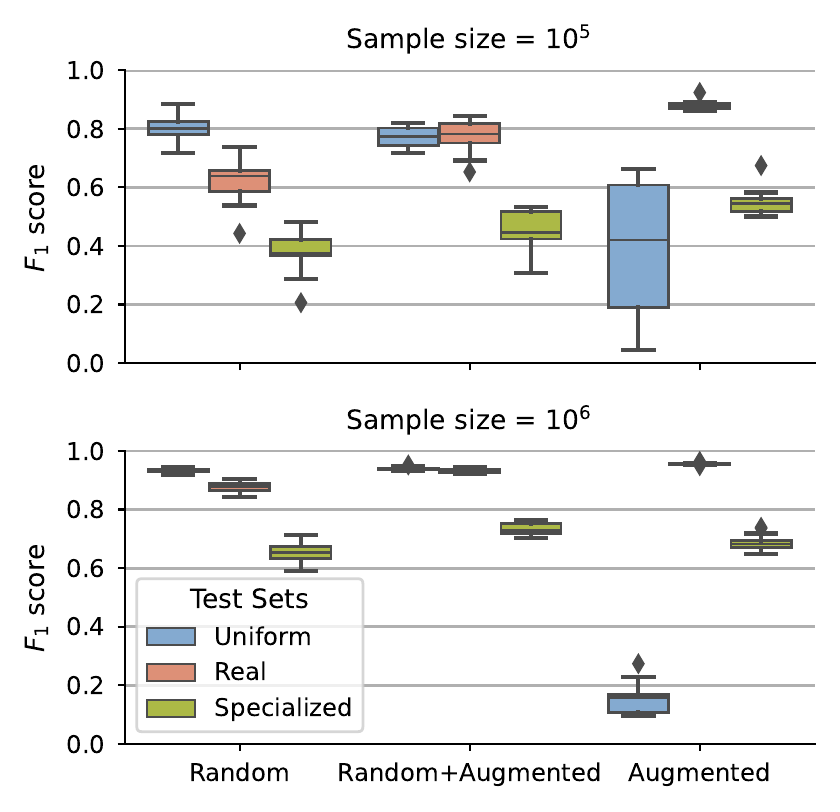}
    \caption{Effect of different sampling approaches with different blends of augmented real designs on the classification performance ($F_1$ score) of Autogluon models evaluated on the uniform, real, and specialized test sets for sample sizes of $10^5$ (top) and $10^6$ (bottom). For each sampling approach, ten independent models are trained with samples drawn at random from the random and/or augmented datasets respectively.}
    \label{fig:realism_performance}
\end{figure}

\subsubsection*{How relevant is realism in a dataset?}
Taking another perspective on the representativeness of a dataset, we look at the question of realism. Figure~\ref{fig:realism} shows the relative distribution of three different sub-sample sets compared to the design space and more importantly to real designs. First, Fig.~\ref{fig:realism} confirms that real designs are indeed clustered, and most of them, are located near the edge of the design space. Consequently, while the diverse sub-set has samples near real designs, their density near real design clusters is low. In contrast, the augmented-only sub-set has a high density near real design, but it has no samples in most areas of the design space. In between, the sub-set combining diverse samples with samples from the augmented dataset display a good balance between design space and real-design cluster coverage. In many real-world applications of machine learning, the model may be tested on new real-world designs. It is quite likely that features of new real-world designs look more similar to other existing real-world designs and not necessarily to synthetic samples randomly generated from a large design space. This may create an issue, as machine learning models trained on the latter may perform poorly on real-world designs, although, they may get very good validation performance. Hence, we propose considering realism as an important factor in both the sampling and testing of datasets.

To confirm the interest in including augmented data, we verify how this translates to the performance of classifiers. For this, we consider train sets composed of (i) random samples, (ii) 70\% random samples and 30\% augmented samples, and (iii) only augmented samples. Figure~\ref{fig:realism_performance} shows the $F_1$ score of the trained classifier for each dataset combination and for all three test sets. It highlights that the combined dataset enables very good performance on both the uniform and the real test sets, with only minor differences compared to their respective reference case. The combined dataset also slightly improves the classification performance on the specialized test set, which is expected as this test set also derives from real designs. Interestingly, Fig.~\ref{fig:realism_performance} also shows that the augmented-only dataset can lead to over-fitting. Indeed, the model trained on $10^6$ samples has seen the whole dataset and has a lower $F_1$ score on the uniform test set than the models trained with smaller subsets of that dataset.

In summary, we highlight the importance of considering ``realism'' as an objective for sampling methods to create machine learning datasets. Adding realism as an objective of dataset sampling methods in machine learning means prioritizing data that is representative of the real-world scenarios that the model is intended to operate in. This can be achieved by collecting data from real-world sources. However, if there is little data available from real-world sources, then it is also important to cover the design space. The objective is to create a dataset that reflects the variability and complexity of the real-world designs as well as all possible variations of a design, helping to ensure that the model is well-suited to real-world applications and can perform well on a diverse range of test cases.





\section{Conclusion}
This paper has discussed the many facets and challenges of synthetic tabular dataset creation in engineering design.
We have provided guidelines for data representation and selection, modeling and simulation, and characterization and verification, all of which are critical steps in ensuring that the generated dataset is accurate, reliable, and representative of the real-world system being modeled or observed.
For each step, we have provided methods and examples to support researchers, using a case study of the centrifugal compressor design dataset. We have highlighted that achieving diversity in design, diversity in performance and \textit{realism} are sometimes competitive objectives, that should and can be factored in when selecting data generation methods. Further, we have illustrated the critical role of well-crafted test sets to ensure a relevant performance assessment of machine learning models.

Overall, the guidelines and considerations presented in this paper provide a valuable resource for researchers and practitioners involved in synthetic tabular dataset creation in engineering design. By following these guidelines, it is possible to create high-quality datasets that accurately reflect the real-world system being modeled or observed, and that are well-suited for use in machine learning and other data-driven applications.

Future work in this area includes the development of standardized procedures for creating and evaluating synthetic datasets, the exploration of new modeling and simulation techniques, and the investigation of ways to reduce bias in synthetic datasets.


\section*{Acknowledgments}
The Swiss National Science Foundation is acknowledged for its financial support (grant P500PT\_206937). The authors also acknowledge the MIT SuperCloud and Lincoln Laboratory Supercomputing Center for providing high-power computing resources that have contributed to the research results reported within this paper.



\bibliographystyle{asmeconf}  
\bibliography{references}

\appendix

\section{Parameters of the compressor model}

Table~\ref{tab:compressor_features} lists the geometrical and operating condition parameters needed to run the mean-line analysis model for centrifugal compressors. In addition to falling in the provided range, the inlet pressure and the temperature are also adjusted depending on the selected working fluid.

\begin{table}[!ht]
\caption{Descriptions of the parameters of the compressor model including acceptable range and units }\label{tab:compressor_features}%
\centering{%
\footnotesize
\begin{tabular}{@{}clrl@{}}
\toprule
\multicolumn{4}{c}{\textit{Independent geometrical parameters}} \\
\midrule
   $r_4$      & Impeller tip radius                    & $[5, 250]$ & \si{\milli\meter}\\
   $\beta_2$  & Mean-line impeller inlet blade angle   & $[-60, 0]$  & \si{\degree} \\
   $\beta_4$  & Mean-line impeller outlet blade angle  & $[-70, -35]$ & \si{\degree} \\
   $e_{b}$    & Mean blade thickness                   & $[0.1, 3]$ & \si{\milli\meter}\\
   $Z_b$      & Number of blades                       & $\llbracket 5, 20 \rrbracket$ & -\\
\midrule
\multicolumn{4}{c}{\textit{Dependent geometrical parameters}} \\
\midrule
   $r_1$    & Inducer inlet radius & $[r_{2s}, r_4]$  & \si{\milli\meter}   \\
   $r_{2h}$ & Impeller hub radius  & $[0.1r_4, 0.5r_4]$ & \si{\milli\meter} \\
   $r_{2s}$ & Impeller shroud radius & $[1.2r_{2h}, 0.8r_4]$ & \si{\milli\meter} \\
   $r_5$  & Diffuser outlet radius & $ [r_4, 4r_4]$ & \si{\milli\meter} \\
   $b4$ & Impeller outlet blade height  &  $[0.015 r_4, 0.3 r_4]$ & \si{\milli\meter} \\
   $b5$ & Diffuser channel width &  $[0.5b_4, 1.5b_4]$ & \si{\milli\meter} \\
   $\beta_{2s}$ & Impeller inlet shroud blade angle  &  $[\beta_2-20, \beta_2]$ & \si{\degree} \\
   $e_{tp}$ & Tip clearance &  $[0.01 b_4, 0.15 b_4]$ & \si{\milli\meter} \\
   $e_{bk}$ & Impeller backface clearance &  $[0.001r_4, 0.15r_4]$ & \si{\milli\meter} \\
   $l_{ind}$ & Inducer length & $[r_4, 4r_4]$ & \si{\milli\meter} \\
   $Z_s$ & \multicolumn{2}{l}{Number of splitter blades \hfill $0 \text{ if } Z_b > 11 \text{ else } Z_b$} & - \\
\midrule
\multicolumn{4}{c}{\textit{Fixed geometrical parameters}} \\
\midrule
   $Ra$ & Inducer/Impeller surface roughness & $\num{1.2e-5}$ & \si{\meter} \\
   $c_{b}$ & Blockage coefficients & $1$  & -  \\
\midrule
\multicolumn{4}{c}{\textit{Independent operating condition parameters}} \\
\midrule
    & \multicolumn{3}{p{7.5cm}@{}}{Working fluid: air, ammonia, isobutane, pentane, propane, R1234yf, R134a, R245fa}   \\
   $Ma_{2,1}$ & Inlet Mach number & $[\SI{5e-2}, 0.7]$  & - \\
   $Ma_{4,1}$ & Impeller tip Mach number & $[\SI{5e-2}, 2.5]$  & - \\
\midrule
\multicolumn{4}{c}{\textit{Dependent operating condition parameters}} \\
\midrule
   $T_{1}$ & Inducer inlet temperature & $[170, 400]$ & \si{\kelvin} \\
   $P_{r,1}$ & Reduced inducer inlet pressure & $[1, 100]$ & - \\
\bottomrule
\end{tabular}
}
\end{table}






\end{document}